\documentclass[11pt]{article}

\usepackage[final]{acl}
\usepackage{float}
\usepackage{times}
\usepackage{latexsym}
\usepackage{pifont}

\usepackage[T1]{fontenc}
\usepackage{booktabs}
\usepackage[utf8]{inputenc}
\usepackage{amsmath}

\usepackage{amssymb}
\usepackage{amsthm}
\usepackage{multirow} 

\usepackage{tabularx}
\usepackage{array}
\newcolumntype{Y}{>{\centering\arraybackslash}X}

\usepackage{amsfonts}
\usepackage{microtype}

\usepackage{inconsolata}

\usepackage{graphicx}

\title{Do LLM Embedding Spaces Recover Expert Structure?}

\author{
Yixuan Zhu\textsuperscript{} \\
Zhongnan University of \\
Economics and Law
\And
Zhenke Duan\textsuperscript{*,\dag} \\
CodeSoul.co \\
\texttt{zhenkeduan03@gmail.com}
\And
Fanghen Li \\
Zhongnan University of \\
Economics and Law \\
CodeSoul.co
}

\begin{document}
\maketitle

\begin{center}
\small
\textsuperscript{*}Equal contribution. \quad
\textsuperscript{\dag}Corresponding author.
\end{center}

\begin{abstract}
Pretrained text embeddings are increasingly used as representational maps, yet high category separability does not imply that their geometry recovers expert-defined structure. We study this problem in mental-health-related language, where symptom relations provide an external reference and online communities introduce strong domain, affective, stylistic, and discourse confounds. Using 28 Reddit communities, we compare pretrained and supervised fine-tuned Qwen3 embedding spaces at two scales (0.6B and 4B). We construct category prototypes, evaluate their representational dissimilarity matrices against an expert symptom matrix with representational similarity analysis, and complement this global test with prototype-based typicality and multi-baseline confound controls. Pretrained embeddings show measurable alignment with expert structure within the mental-health subset; fine-tuning strengthens this alignment most at the finest category level; and larger scale improves both zero-shot alignment and supervision-induced gains. Residual alignment remains substantial after controlling for VAD, LIWC, lexical style, and topic-distribution structure. These results suggest that LLM embeddings can recover expert-relevant category geometry, but this recovery is level-dependent and should be tested against explicit confounds rather than inferred from classification alone.

\end{abstract}

\section{Introduction}

Text embeddings from large language models are increasingly used not only as downstream features but as representational maps: distances among embedded texts can be compared with independently specified conceptual structures. This shift raises a question for structure-level evaluation: when embeddings separate categories, do they recover fine-grained conceptual organization, or do they mainly reflect coarse distributional cues? RSA-style analyses provide a way to compare model geometry with external reference structures, but matrix-level correlations can also obscure which aspects of the geometry drive the observed alignment \citep{kondapaneni2025rsvc,shah2024cognitiveintelligence}.

\begin{figure}[t]
  \centering
  \includegraphics[width=\linewidth]{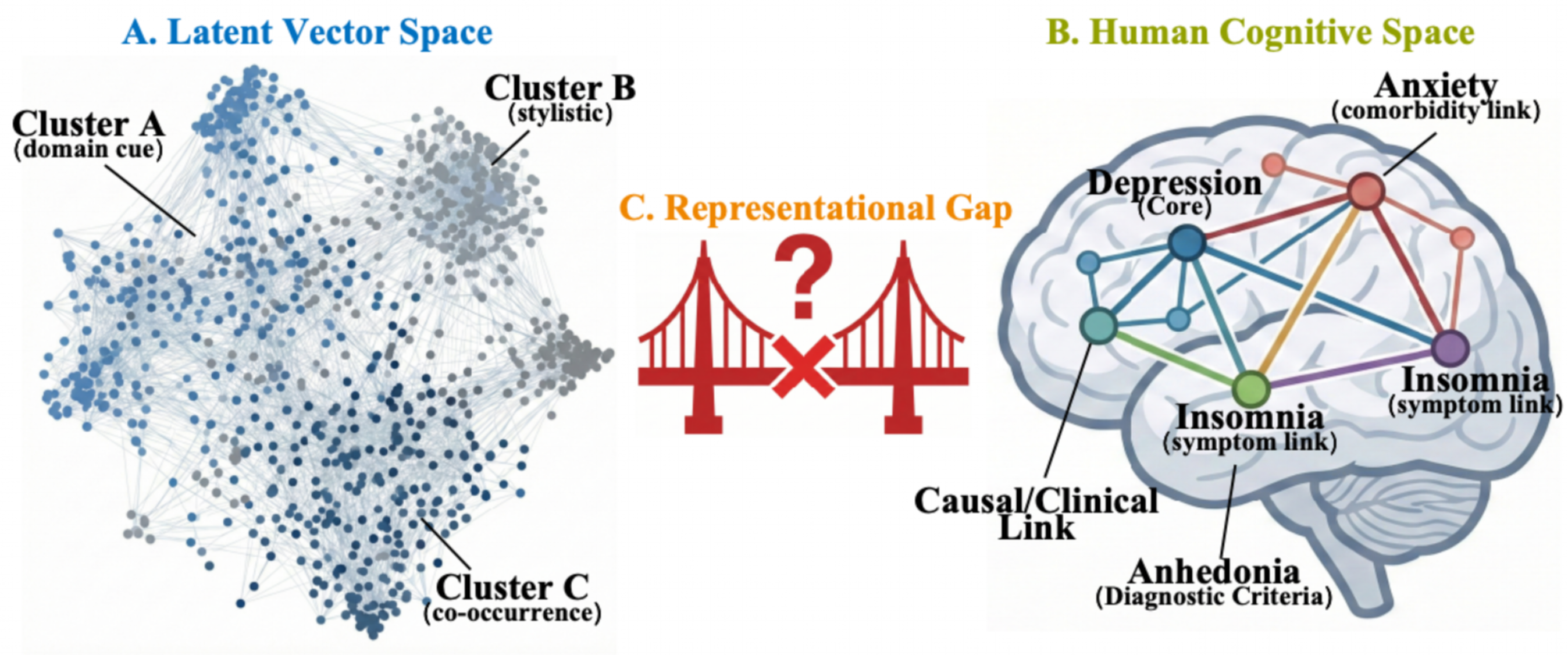}
\caption{\textbf{The Representational Gap in Mental-Health Embeddings.} (A) Embedding spaces may cluster communities by non-clinical cues such as subreddit jargon, style, and broad domain separation. (B) Expert structure organizes categories by symptom overlap and graded boundaries. (C) We test whether model geometry aligns with this expert structure beyond domain and other non-clinical cues.}
  \label{fig:rep_gap}
\end{figure}

Mental-health language makes this problem sharp. Mental-health-related categories exhibit symptom overlap and graded boundaries, yielding an external expert reference for fine-grained category relations. Yet online communities also contain strong non-clinical signals, including subreddit jargon, stylistic norms, and a broad mental-health/control split, that may dominate the learned space. A model may therefore classify communities or cluster them cleanly without encoding the symptom-level relations that motivate the category structure. Figure~\ref{fig:rep_gap} illustrates this gap between community separability and expert-aligned geometry.

We ask whether embedding spaces recover expert-defined symptom structure beyond coarse domain separation, and how task-specific supervision changes this geometry. Using 28 online communities (17 mental-health, 11 controls), we compare pretrained and supervised fine-tuned embedding spaces. This framing shifts evaluation from community classification to the category structure encoded in the representation space, consistent with evidence that language models can recover meaningful similarity structure but require explicit controls for confounds \citep{marjieh2024sensory,ogg2024turingrsa}. We focus on three questions: (1) whether embeddings align with expert-defined symptom structure beyond domain separation; (2) whether supervision reshapes global alignment, local boundary organization, or both; and(3) whether the observed geometry reflects expert-relevant relations rather than discourse, affective,or stylistic regularities.

To answer these questions, we combine three analyses: representational similarity analysis (RSA) for global alignment with expert symptom structure; prototype-based typicality for within-category cohesion and boundary confusions \citep{vemuri2024typicalityeffect}; and multi-baseline confound controls for affective valence--arousal--dominance (VAD), Linguistic Inquiry and Word Count (LIWC), stylistic, and topic-based regularities.Our contributions are as follows:
\begin{itemize}
\item We present a representation-level framework for testing whether embedding spaces recover expert-defined structure beyond domain separation, combining RSA, prototype-based typicality, and multi-baseline confound controls.
\item We show a level-dependent pattern of expert-structure recovery: pretrained embeddings encode symptom-level organization within the mental-health subset, and supervision strengthens this alignment most at the finest category level.
\item We compare two model scales (0.6B and 4B), showing that larger scale improves both zero-shot alignment and fine-tuning gains, while supervision and scale jointly reduce associations with non-clinical baselines and yield more category-specific geometry.
\end{itemize}

\section{Related Work}

\subsection{Embedding geometry and structure-level evaluation}

Text embeddings are increasingly analyzed not only as features for downstream tasks, but as geometric spaces whose distances, directions, and neighborhoods can be compared with human conceptual structure. Prior work shows that such geometry can recover interpretable dimensions of conceptual knowledge \citep{Grand_2022_04,Xu_2025_06}. In NLP, however, this structure is often operationalized through scalar similarities, such as cosine similarity between pooled sentence representations in models like SBERT \citep{reimers2019sentence}. These measures are useful, but they compress high-dimensional geometry into pairwise closeness and provide limited information about how categories are organized as a relational system.

Representational similarity analysis (RSA) provides a structure-level alternative. By comparing model-derived representational dissimilarity matrices with external relational references, RSA tests whether two systems organize the same items in similar ways, making it well suited for evaluating alignment between model geometry and human- or expert-defined structures \citep{Fairhall_2024_03}. Yet global matrix correlations can hide local failures: two spaces may show similar overall structure while differing around category boundaries. Methods that retain finer pairwise, interaction-level, or topological information capture complementary aspects of geometry, but they do not by themselves specify how category-level representations should be evaluated against an external expert reference \citep{khattab2020colbert,Brown_2025_06}. Our work combines these views: we use RSA to test global expert alignment and prototype-based analyses to examine where that alignment sharpens or breaks down locally.

This framing reflects cautions against treating embedding geometry as a complete account of meaning. Model--human alignment is partial and feature-dependent, with weaker recovery of attributes such as sensorimotor features \citep{Xu_2025_06}. Different representation sources can induce distinct relational patterns, suggesting that embedding spaces reflect multiple organizing factors rather than a single semantic axis \citep{Carota_2023_07}. We therefore test whether category-level geometry reflects an externally specified symptom structure rather than competing explanations such as domain membership, affective similarity, or other non-clinical regularities.

\subsection{Interpretability and local structure in embedding spaces}

Embedding interpretability is often approached through probing, attribution, token-level alignment, and subspace-based methods \citep{opitz2025survey}. Probing tests whether linguistic or semantic information is recoverable from embeddings \citep{conneau2018cram,muennighoff2023mteb}, but recoverability does not determine the topology or neighborhood structure of a space \citep{belinkov2022probing}. Attribution and token-level alignment explain prediction-relevant inputs or fine-grained contextual correspondences \citep{sundararajan2017axiomatic,han2020influence,khattab2020colbert,zhang2020bertscore}; subspace methods make selected semantic dimensions more interpretable \citep{opitz2022sbert}. These approaches are valuable, but they do not directly test whether category-level geometry aligns with an external expert reference.

Our setting requires this representation-level test. We examine whether category prototypes, similarity relations, and boundary regions form a structured space consistent with expert-defined symptom relations. Prototype distances quantify within-category cohesion and boundary-prone instances, while topic-based probes identify lexical fields associated with boundary blurring or sharpening. Together, these analyses complement global RSA by showing how local regions of the embedding space are organized, rather than serving as standalone explanations of individual predictions.

\subsection{Mental-health language as a structured but confounded testbed}

Mental-health language is a useful testbed for expert-aligned representation analysis because it is both structured and confounded. Mental-health-related categories involve symptom overlap, graded boundaries, and clinically meaningful distinctions, making them suitable for evaluating fine-grained category organization. Prior work reports systematic semantic, syntactic, and referential variation in psychiatric and mental-health language, with different computational representations capturing different aspects of this variation \citep{He_2024_03,Palominos_2024_12}.

At the same time, online communities introduce strong non-clinical confounds. Similarity between subreddits may reflect lexical overlap, discourse conventions, community-specific jargon, or style rather than expert-relevant relations \citep{nikolaev2023semantic}. Categories may also appear close because they share affective profiles in valence, arousal, or dominance, not because they share symptom-level structure \citep{marjieh2024sensory}. Because these factors are intertwined with the relational structure of interest, they cannot simply be removed. This setting therefore allows us to ask whether embedding geometry remains aligned with an expert reference after controlling for coarse domain separation and affective, stylistic, and discourse-level confounds.

\section{Method}

\paragraph{Overview.}
We encode each document with a pretrained encoder, optionally fine-tune the encoder on community labels, and analyze the resulting category-level geometry in both zero-shot and fine-tuned spaces. Our primary analyses compare model-derived category structure with expert-defined and domain-control reference structures using representational similarity analysis (RSA). We then use prototype-based typicality, boundary ambiguity, and topic-based probes to interpret local cohesion, boundary confusion, and similarity patterns in the learned spaces. Appendix~\ref{app:pipeline} provides a schematic overview of the full analysis pipeline.

\paragraph{Embedding spaces.}
Let $\mathcal{D}=\{(x_i,y_i)\}_{i=1}^{N}$ denote documents $x_i$ with category labels $y_i \in \{1,\dots,C\}$, where $C=28$ consists of 17 mental-health categories and 11 domain-control communities. A pretrained encoder $f_{\theta_0}$ maps each document to a $d$-dimensional zero-shot embedding,
$\mathbf{h}_i^{\mathrm{ZS}} = f_{\theta_0}(x_i)$.
For the fine-tuned setting, we optimize the encoder with a linear classifier under cross-entropy loss and use the resulting encoder $f_{\theta^\star}$ to obtain
$\mathbf{h}_i^{\mathrm{FT}} = f_{\theta^\star}(x_i)$
(Appendix~\ref{app:embedding_spaces}). All downstream analyses are performed in both representation settings.

\paragraph{Category-level representational geometry and reference alignment.}
For each category $c$, we construct a prototype by averaging document embeddings:
\begin{equation}
  \boldsymbol{\mu}_c = \frac{1}{|\mathcal{I}_c|}\sum_{i\in\mathcal{I}_c}\mathbf{h}_i,
  \qquad
  \mathcal{I}_c=\{i:\,y_i=c\}.
\end{equation}
The category prototypes define a representational dissimilarity matrix (RDM),
\begin{equation}
  \mathrm{RDM}_{\mathrm{LLM}}(c,c') = d(\boldsymbol{\mu}_c,\boldsymbol{\mu}_{c'}),
\end{equation}
where $d$ denotes cosine distance. We compare $\mathrm{RDM}_{\mathrm{LLM}}$ with two reference matrices: an expert-derived symptom matrix $\mathrm{RDM}_{\mathrm{Expert}}$ and a domain-control matrix $\mathrm{RDM}_{\mathrm{Domain}}$. For any two RDMs $A, B \in \mathbb{R}^{C \times C}$, RSA is defined as the Spearman rank correlation between their upper-triangular entries, excluding the diagonal:
\begin{equation}
  \mathrm{RSA}(A, B) \;=\; \rho_{\mathrm{Spearman}}\!\bigl(\mathrm{vec}_{\triangle}(A),\, \mathrm{vec}_{\triangle}(B)\bigr).
  \label{eq:rsa_def}
\end{equation}
We assess significance by permuting category labels and estimate uncertainty with bootstrap resampling over documents (Appendix~\ref{app:rdm}).

\paragraph{Prototype-based typicality and boundary analysis.}
We analyze local structure through document typicality, prototype margins, and expert-induced boundary ambiguity. The \emph{prototype-based typicality distance} of document $x_i$ is
\begin{equation}
  \tau_i = d(\mathbf{h}_i,\boldsymbol{\mu}_{y_i}),
\end{equation}
where smaller values indicate greater centrality within the labeled category. Category-level cohesion is summarized as
$\mathrm{Coh}(c) = \frac{1}{|\mathcal{I}_c|}\sum_{i\in\mathcal{I}_c}\tau_i$.
The \emph{prototype margin} of document $x_i$ is
\begin{equation}
  m_i = \min_{c' \neq y_i} d(\mathbf{h}_i,\boldsymbol{\mu}_{c'}) - d(\mathbf{h}_i,\boldsymbol{\mu}_{y_i}),
\end{equation}
with $m_i < 0$ indicating a boundary-confusion instance. We further define the \emph{boundary alignment score}
\begin{equation}
  \Delta_{\mathrm{align}} = \mathrm{Amb}_{+} - \mathrm{Amb}_{-},
\end{equation}
where $\mathrm{Amb}_{+}$ and $\mathrm{Amb}_{-}$ denote mean boundary ambiguity over expert-neighbor and non-neighbor category pairs, respectively. Larger $\Delta_{\mathrm{align}}$ indicates that residual boundary uncertainty is concentrated among category pairs that are nearby in expert structure. Appendix~\ref{app:boundary} gives the full definitions of pairwise ambiguity and the expert-neighborhood graph.

\paragraph{Idealized analysis under a class-conditional Gaussian model.}
To clarify the geometric meaning of these measures, we consider a class-conditional Gaussian model in which each category $c$ has latent mean $\boldsymbol{\mu}_c^\star$ and isotropic covariance $\sigma_c^2 I_d$:
\begin{equation}
  \mathbf{h} \mid y=c \;\sim\; \mathcal{N}(\boldsymbol{\mu}_c^\star,\,\sigma_c^2 I_d).
\end{equation}
Under this model, we obtain two results that connect prototype geometry to cohesion and boundary confusion (proofs in Appendix~\ref{app:proofs}).

\textbf{Proposition 1} (Typicality and within-category variance).
\textit{For any category $c$,}
\begin{equation}
  \mathbb{E}\big[\|\mathbf{h}-\boldsymbol{\mu}_c^\star\|^2 \mid y=c\big] = d\,\sigma_c^2.
\end{equation}
\textit{Thus, expected squared typicality is proportional to within-category variance; smaller typicality corresponds to tighter cohesion.}

\textbf{Proposition 2} (Boundary confusion and prototype separation).
\textit{Under two-class isotropic Gaussians with common variance $\sigma^2 I_d$ and prototype separation $\Delta_{c,c'}=\|\boldsymbol{\mu}_c^\star - \boldsymbol{\mu}_{c'}^\star\|$, the probability that a sample from category $c$ falls on the wrong side of the prototype boundary is}
\begin{equation}
  \mathbb{P}\!\left(S(\mathbf{h};c,c')<0 \mid y=c\right)
  = \Phi\!\left(-\frac{\Delta_{c,c'}}{2\sigma}\right),
\end{equation}
\textit{where $\Phi$ is the standard normal CDF. Boundary confusion decreases as prototype separation increases and increases as within-category variance grows.}

\textbf{Corollary 1} (Fine-tuning increases boundary alignment).
\textit{Suppose supervision increases $\Delta_{c,c'}/\sigma$ for non-neighbor pairs while not substantially decreasing it for expert-neighbor pairs. Then, by Proposition~2, boundary confusion becomes more concentrated on expert-neighbor pairs, implying}
\begin{equation}
  \Delta_{\mathrm{align}}^{\mathrm{FT}} > \Delta_{\mathrm{align}}^{\mathrm{ZS}}.
\end{equation}

\begin{figure}[t]
  \centering
  \includegraphics[width=\linewidth]{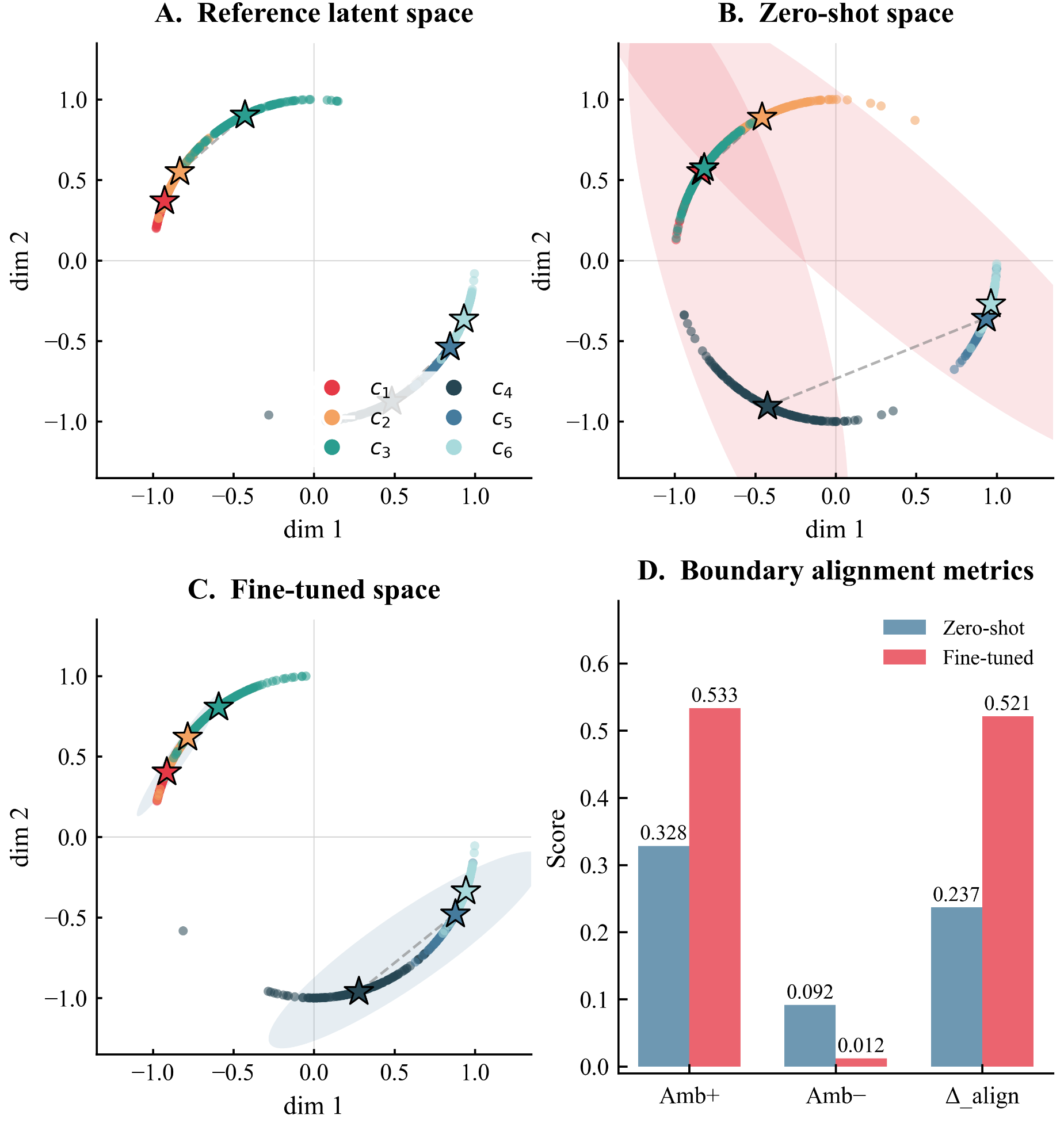}
\caption{\textbf{Synthetic boundary realignment.}
(A) Reference latent space induced by a synthetic expert-neighborhood graph. (B) Zero-shot space after coarse domain distortion and increased within-category variance, producing diffuse boundary ambiguity. (C) Fine-tuned space after structured supervision, where off-structure ambiguity is reduced and residual ambiguity concentrates among expert-neighbor pairs. (D) Summary metrics: fine-tuning increases ambiguity among expert-neighbor pairs ($\mathrm{Amb}_{+}$), decreases ambiguity among non-neighbor pairs ($\mathrm{Amb}_{-}$), and increases the boundary alignment score $\Delta_{\mathrm{align}}$.}
  \label{fig:synthetic_boundary}
\end{figure}

\paragraph{Synthetic boundary realignment experiment.}
To illustrate Corollary~1, we construct a six-category, two-dimensional Gaussian setting with an explicit expert-neighborhood graph (Figure~\ref{fig:synthetic_boundary}). Adding coarse domain distortion and extra within-category noise to the reference space produces diffuse boundary ambiguity in the zero-shot setting. After simulated fine-tuning, off-structure ambiguity is suppressed while residual ambiguity remains concentrated among expert-neighbor pairs. Quantitatively, the boundary alignment score rises from $0.237$ to $0.521$, and mean category cohesion tightens from $1.160$ to $0.979$. The effect of supervision is therefore not to eliminate all boundaries uniformly, but to reduce off-structure ambiguity and concentrate residual uncertainty where expert structure predicts it.

\paragraph{Topic-based interpretability probes.}
To interpret local boundary structure, we fit an Embedded Topic Model (ETM) in each representation setting to estimate document-topic proportions and topic-word distributions. We summarize topics by their highest-probability keywords and assign each topic to the category in which it is most prevalent. This provides a topic-level account of cross-category proximity and boundary-related lexical overlap. The ETM objective is given in Appendix~\ref{app:etm}.

\section{Experiments}
\subsection{Dataset and Representation Settings}

We use the \textit{Reddit Mental Health Dataset}, which contains posts and text-derived features from 28 subreddits spanning 2018--2020 and was curated to study language change in mental-health communities before and during the early COVID-19 period \citep{low2020natural}. The dataset includes 15 disorder- or support-specific mental-health subreddits (e.g., r/depression, r/anxiety, r/suicidewatch), 2 broad mental-health forums (r/mentalhealth, r/COVID19\_support), and 11 non-mental-health control subreddits (e.g., r/legaladvice, r/personalfinance, r/jokes, r/fitness).

We treat each subreddit as a category label ($C=28$) and analyze category-level geometry over mental-health and control communities. To reduce subreddit-size imbalance, we construct balanced analysis sets by stratified subsampling and estimate uncertainty with bootstrap resampling, following the dataset authors' recommendation. This dataset is well suited to our question because it combines expert-relevant symptom relations with strong contextual signals, including subreddit-specific jargon, discourse conventions, and a broad mental-health/control distinction. It allows us to test whether learned geometry reflects expert-defined category structure rather than category separability or coarse domain cues alone.

\subsection{Embeddings and Topic Modeling}
\subsubsection{Zero-shot and fine-tuned embedding spaces}
We compare two representation settings at two model scales: pretrained zero-shot spaces and supervised fine-tuned spaces for Qwen3-Embedding-0.6B and Qwen3-Embedding-4B. In the fine-tuned setting, each model is optimized on the 28-community classification objective with a linear head and cross-entropy loss. We then discard the classification head and use the encoder outputs for all downstream geometry analyses, including category prototypes, RDM estimation, RSA, and local boundary analysis.

Following the model specification, documents are represented as fixed-length 1024-dimensional vectors for Qwen3-Embedding-0.6B. The training curve is reported in Appendix~\ref{app:training_curves}: accuracy rises rapidly and stabilizes near $\sim$0.83, while loss remains low with moderate batch-level variation. Qwen3-Embedding-4B follows a similar trajectory and reaches a higher accuracy of $\sim$0.87, consistent with its larger representational capacity. We treat these fine-tuned models not as classification endpoints, but as task-conditioned embedding spaces that can be compared with their zero-shot counterparts and across scales to examine how supervision and model size reshape category geometry.

\subsubsection{Topic-based interpretability probe}
As an auxiliary interpretability probe, we estimate topic structure in each representation setting with an Embedded Topic Model (ETM) and compare it with bag-of-words LDA. Table~\ref{tab:topic_metrics} reports topic quality metrics (TD, iRBO, NPMI, and exclusivity) and label-alignment metrics for topic-induced clusters (ARI and NMI). Fine-tuning improves topic structure across metrics: topic diversity and exclusivity increase, NPMI becomes less negative, and topic clusters align more closely with subreddit labels. LDA serves as a conservative lexical reference.

\begin{table}[H]
\centering
\caption{\textbf{Topic quality and label alignment.} ETM results for zero-shot 
vs fine-tuned embeddings across model scales, with an LDA baseline.}
\label{tab:topic_metrics}
\setlength{\tabcolsep}{5pt}
\renewcommand{\arraystretch}{1.05}
\resizebox{\columnwidth}{!}{%
\begin{tabular}{llcccccc}
\hline
\textbf{Model} & \textbf{Setting} & \textbf{TD} & \textbf{iRBO} & 
\textbf{NPMI} & \textbf{Excl} & \textbf{ARI} & \textbf{NMI} \\
\hline
— & LDA & 0.112 & 0.336 & -0.308 & 0.104 & 0.176 & 0.421 \\
\multirow{2}{*}{0.6B} 
& Zero-shot  & 0.293 & 0.342 & -0.182 & 0.203 & 0.230 & 0.448 \\
& Fine-tuned & 0.347 & 0.418 & -0.057 & 0.354 & 0.319 & 0.580 \\
\multirow{2}{*}{4B}   
& Zero-shot  & 0.318 & 0.371 & -0.134 & 0.247 & 0.271 & 0.487 \\
& Fine-tuned & 0.389 & 0.453 & -0.021 & 0.412 & 0.368 & 0.634 \\
\hline
\end{tabular}%
}
\end{table}

\subsection{Construction of the expert reference structure}

\begin{figure}[t]
  \centering
  \includegraphics[width=\linewidth]{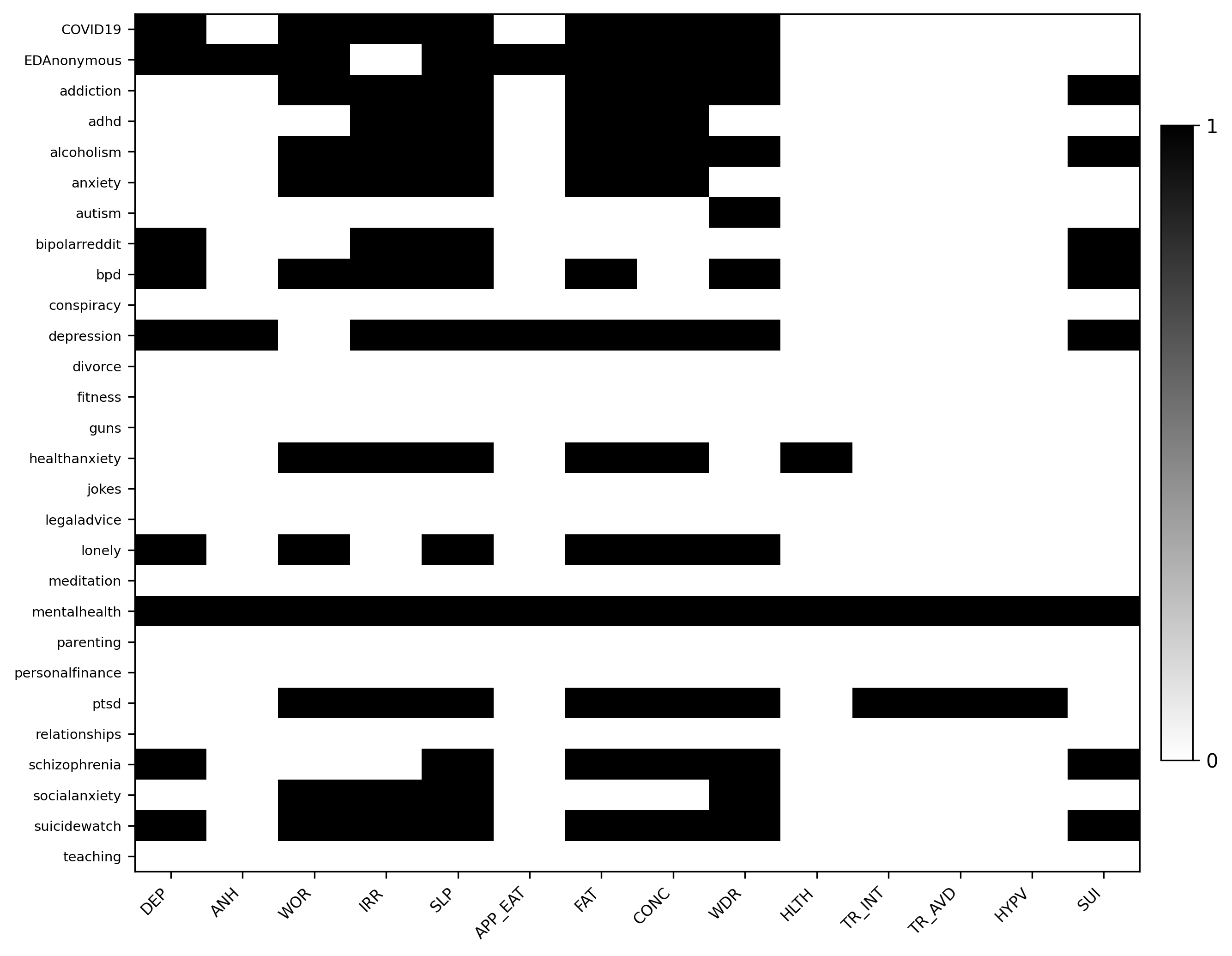}
  \caption{\textbf{Binary expert reference matrix.} Rows correspond to subreddit categories and columns to symptom dimensions in $\mathcal{M}$. Black cells indicate presence (1) and white cells indicate absence (0). Control communities are assigned the all-zero vector by construction.}
  \label{fig:expert_symptom_bw}
\end{figure}

Our alignment analyses require an external reference over the same 28 categories. We construct this reference from a symptom vocabulary $\mathcal{M}$, consisting of expert-informed dimensions that capture broad symptom and behavioral features of the mental-health-related communities. Each category $c$ is encoded as a binary symptom-profile vector $\mathbf{s}_c \in \{0,1\}^{|\mathcal{M}|}$, where each entry indicates the presence or absence of a symptom dimension. Collecting these vectors yields a binary symptom matrix $S \in \{0,1\}^{C \times |\mathcal{M}|}$, shown in Figure~\ref{fig:expert_symptom_bw}, with rows corresponding to subreddit categories and columns to symptom dimensions. Control communities are assigned the all-zero vector by construction, reflecting the absence of targeted mental-health dimensions in the reference scheme. We derive the expert representational dissimilarity matrix (RDM) by computing pairwise Jaccard distances between symptom profiles, $\mathrm{RDM}_{\mathrm{Expert}}(c,c') = d_J(\mathbf{s}_c,\mathbf{s}_{c'})$, where $d_J$ denotes Jaccard distance on binary symptom sets.

The all-zero profiles assigned to control communities encode their role as a structured negative contrast, rather than an alternative expert-defined symptom profile. Under this design, the expert RDM tests a targeted hypothesis: mental-health categories are organized by shared symptom dimensions, while control categories lie outside this symptom vocabulary by definition. This makes the reference intentionally asymmetric, targeting symptom-level structure within and around the mental-health subset rather than a universal taxonomy over all community types. For transparency, Figure~\ref{fig:expert_symptom_bw} reports the full binary matrix; Appendix~\ref{app:expert_reference} provides the symptom dimensions, coding criteria, and category assignments; and Appendix~\ref{app:annotation_protocol} reports the annotation protocol and inter-rater reliability statistics, including overall Cohen's $\kappa = 0.81$.

\subsection{Global alignment with expert and domain reference structures}

\begin{figure}[t]
  \centering
  \includegraphics[width=\linewidth, height=0.32\textheight, keepaspectratio]{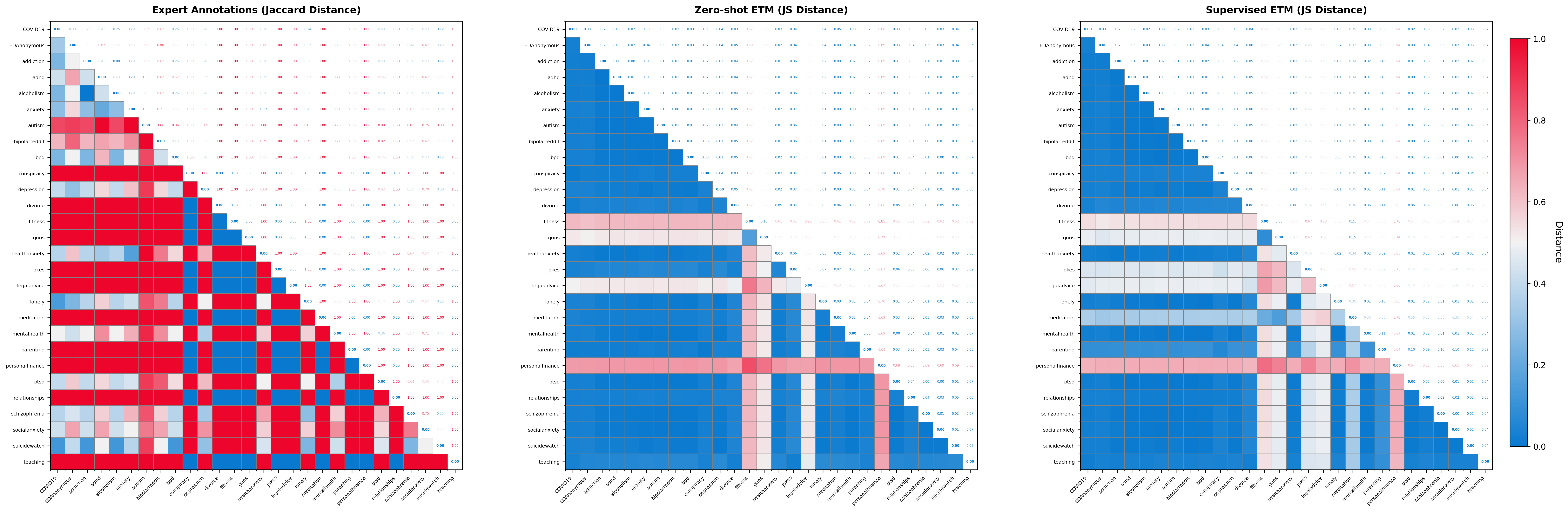}
  \caption{\textbf{Three-way comparison of  representational geometry.}
  Left: expert reference RDM (Jaccard distance). Middle: zero-shot ETM RDM (JS distance). Right: fine-tuned ETM RDM  (JS distance).}
  \label{fig:three_way_rdm}
\end{figure}

We first evaluate whether the category geometry induced by the real data aligns with the expert reference beyond coarse domain separation. For each representation setting, we summarize model geometry with a representational dissimilarity matrix (RDM). Specifically, we compute a prototype for each subreddit category by averaging its document representations and define $\mathrm{RDM}_{\mathrm{Model}}(c,c') = d(\boldsymbol{\mu}_c,\boldsymbol{\mu}_{c'})$, where $d$ is cosine distance for embedding prototypes and Jensen--Shannon (JS) distance for topic-distribution prototypes (Figure~\ref{fig:three_way_rdm}).

We quantify alignment using RSA (Eq.~\ref{eq:rsa_def}) between $\mathrm{RDM}_{\mathrm{Model}}$ and $\mathrm{RDM}_{\mathrm{Expert}}$. Because the expert and domain RDMs share substantial block-level structure on the full category set (Table~\ref{tab:rdm_collinearity}), we report three complementary analyses: (i) MH-only RSA over the 17 mental-health categories, where the domain confound is absent by construction and which we treat as the \textbf{primary} result; (ii) full-set RSA over all 28 categories; and (iii) partial RSA over the full set, controlling for a domain RDM that encodes MH versus control membership. Uncertainty is estimated by bootstrap resampling over documents when forming prototypes, and significance is evaluated with category-level permutation tests that jointly permute RDM row and column labels. Results are summarized in Table~\ref{tab:rsa_alignment}.

\paragraph{Reference RDM collinearity.}
Before interpreting RSA values, we assess pairwise collinearity among the reference RDMs (Table~\ref{tab:rdm_collinearity}). On the full 28 × 28 category set, the expert and domain RDMs are strongly correlated ($\rho= 0.71$), reflecting the all-zero symptom profiles assigned to control communities. Thus, full-set RSA may conflate fine-grained symptom-level alignment with coarse MH-versus-control separation.
We therefore treat the \textbf{MH-only $17 \times 17$ analysis as the primary RSA result}, since the domain RDM is uninformative within this subset, and interpret the full-set and partial-RSA results as complementary evidence. The VAD reference is only weakly correlated with the expert ($\rho \le 0.21$) and domain ($\rho = 0.23$) RDMs, consistent with the multi-baseline confound analysis in Section~\ref{sec:affective_control}.

On the MH-only subset, pretrained embeddings already show substantial alignment with the expert reference ($\rho = 0.541$ at $0.6$B, $0.628$ at $4$B). Supervision produces the largest gains in this setting ($\Delta\rho = +0.171$ at $0.6$B, $+0.135$ at $4$B), exceeding the gains observed in the full-set ($\Delta\rho \le +0.055$) and partial-RSA ($\Delta\rho \le +0.084$) analyses. These results indicate that fine-tuning most strongly amplifies expert-aligned structure at the mental-health category level, while full-set results should be interpreted in light of the expert--domain collinearity noted above.

\begin{table}[t]
\centering
\small
\caption{Spearman correlation between reference RDMs. 
$^{\dagger}$The domain RDM is constant on the MH-only subset 
and therefore undefined for correlation.}
\begin{tabular}{lcc}
\toprule
\textbf{RDM pair} & \textbf{Full $28\times28$} & \textbf{MH-only $17\times17$} \\
\midrule
Expert vs.\ Domain & 0.71 & n/a$^{\dagger}$ \\
Expert vs.\ VAD    & 0.18 & 0.21 \\
Domain vs.\ VAD    & 0.23 & n/a$^{\dagger}$ \\
\bottomrule
\end{tabular}
\label{tab:rdm_collinearity}
\end{table}

\begin{table}[t]
\centering
\small
\caption{RSA alignment between model and expert RDMs. Spearman $\rho$ is computed between each model-derived RDM and the expert RDM across two model scales ($0.6$B and $4$B) and two representation settings (zero-shot and supervised fine-tuned). The full $28\times28$analysis includes all categories; the MH-only $17\times17$ analysis, our primary result, is restricted to mental-health categories; partial RSA controls for the domain RDM. $95\%$ confidence intervals are estimated by bootstrap resampling over documents. }
\label{tab:rsa_alignment}
\resizebox{\columnwidth}{!}{
\begin{tabular}{llcc}
\toprule
\textbf{Analysis} & \textbf{Setting} & $\boldsymbol{\rho}$ & \textbf{95\% CI} \\
\midrule
\multirow{4}{*}{Full ($28\times28$)}
 & 0.6B Zero-shot   & 0.703 & [0.658, 0.741] \\
 & 0.6B Supervised  & \textbf{0.758} & [0.728, 0.781] \\
 & 4B Zero-shot     & 0.735 & [0.695, 0.766] \\
 & 4B Supervised    & \textbf{0.789} & [0.762, 0.812] \\
\midrule
\multirow{4}{*}{MH-only ($17\times17$)}
 & 0.6B Zero-shot   & 0.541 & [0.404, 0.658] \\
 & 0.6B Supervised  & \textbf{0.712} & [0.617, 0.791] \\
 & 4B Zero-shot     & 0.628 & [0.506, 0.731] \\
 & 4B Supervised    & \textbf{0.763} & [0.675, 0.834] \\
\midrule
\multirow{4}{*}{Partial ($28\times28$)}
 & 0.6B Zero-shot   & 0.567 & [0.481, 0.638] \\
 & 0.6B Supervised  & \textbf{0.651} & [0.581, 0.706] \\
 & 4B Zero-shot     & 0.609 & [0.526, 0.679] \\
 & 4B Supervised    & \textbf{0.682} & [0.615, 0.741] \\
\bottomrule
\end{tabular}
}
\end{table}

\subsection{Local boundary organization and interpretability}

Beyond global RSA, we examine local category organization through prototype-based typicality and topic-based interpretability probes. We ask which categories are internally cohesive, which exhibit diffuse boundary regions, and whether ambiguity is concentrated along expert-neighbor relations. For each document, typicality is defined as its distance to the prototype of its labeled category. Larger distances indicate boundary-prone instances, as these documents lie farther from their category center and closer to competing prototypes.

\begin{figure}[t]
  \centering
  \includegraphics[width=\linewidth]{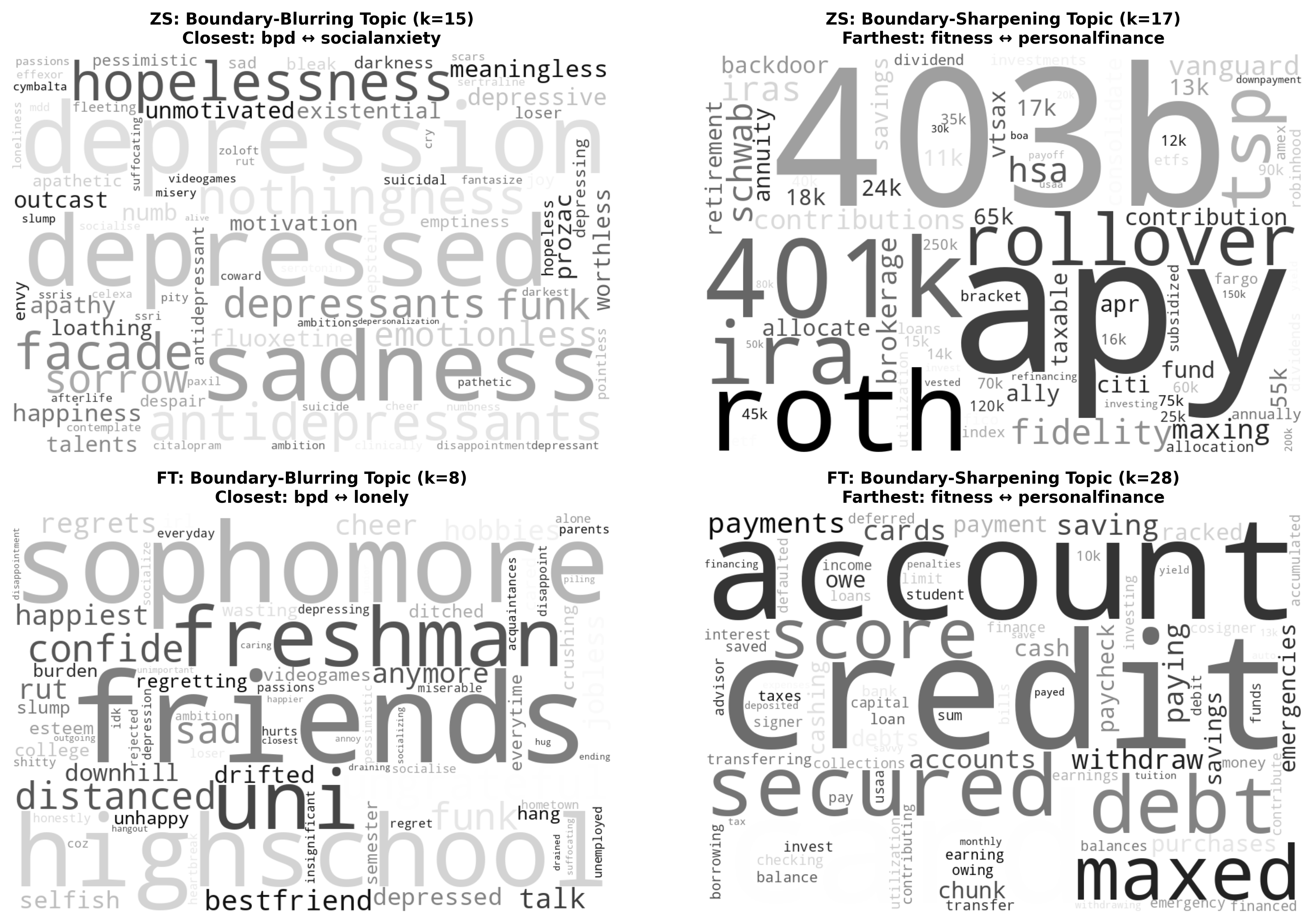}
    \caption{\textbf{Topic-based interpretability of local boundary organization.  (ZS vs.\ FT).}
Representative ETM topics associated with boundary-blurring and boundary-sharpening category pairs in zero-shot (ZS) and fine-tuned (FT) spaces. Boundary-blurring topics reflect shared linguistic content across nearby categories, whereas boundary-sharpening topics show more concentrated category-specific lexical fields. }
  \label{fig:topic_wordcloud}
\end{figure}

To link boundary regions to interpretable linguistic patterns, we examine ETM topics associated with boundary-blurring and boundary-sharpening category pairs in zero-shot and fine-tuned representations for the 0.6B model (Figure~\ref{fig:topic_wordcloud}). Boundary-blurring topics emphasize shared affective or situational narratives, whereas boundary-sharpening topics contain concentrated domain-specific lexical fields, consistent with clearer separations among control communities. The 4B model shows similar patterns, with finer topic distinctions aligned with its stronger global RSA alignment. Overall, supervised fine-tuning reshapes the pretrained space by reducing off-structure ambiguity and concentrating residual boundary uncertainty among expert-neighbor pairs, rather than uniformly increasing separability across categories.

\subsection{Confound Analyses: Affective, Discourse, and 
Stylistic Baselines}
\label{sec:affective_control}

We next test whether model--expert alignment can be explained by non-clinical regularities. Rather than relying on an affective baseline alone, we evaluate four reference structures capturing increasingly content-rich confounds: 
(i) \textbf{VAD} (valence, arousal, dominance), computed from the NRC VAD lexicon as a coarse affective signal \citep{mohammad2018vad}; 
(ii) \textbf{LIWC} affect, social, and cognitive category proportions \citep{boyd2022liwc}; (iii) document-level \textbf{lexical-style} features, including sentence length, type-token ratio, and function-word frequency, following stylometric practice \citep{stamatatos2009survey}; and (iv) the \textbf{ETM topic-distribution RDM} as a discourse-content baseline. For each baseline, we report $\rho(\text{Model},\text{Confound})$ and the partial RSA between model and expert RDMs after controlling for it.

Table~\ref{tab:confound_full} shows two main patterns. First, VAD is not a strong confound: the model--VAD association is weak ($\rho \le 0.20$), and controlling for it changes expert alignment by less than $0.02$. VAD therefore serves as a coarse affective check, indicating that broad affect does not drive the observed geometry, but it is not the strongest nuisance signal. Second, more content-rich baselines explain a larger share of the alignment but do not eliminate it: LIWC affect+social reduces $\rho$ by $0.16$, and ETM topic structure reduces it by $0.29$. Even under the strongest discourse-content control, residual expert alignment remains substantial and significant, indicating that fine-grained category geometry is not reducible to topic structure alone. Across all four baselines, fine-tuning further weakens confound associations; for example, model--LIWC drops from $0.46$ ZS to $0.43$ FT at $0.6$B and from $0.43$ ZS to $0.41$ FT at $4$B. Larger model scale further weakens these associations, suggesting that supervision and scale jointly shift the representation space toward more category-specific organization.

\begin{table}[t]
\centering
\small
\caption{Multi-baseline confound analysis. . Columns under $\rho(\mathrm{M},\mathrm{C})$  report model--confound associations; columns under $\rho(\mathrm{M},\mathrm{E}\!\mid\!\mathrm{C})$  report partial RSA between model geometry and the expert RDM after controlling for each confound. Bold indicates the strongest model--confound association and the largest reduction in expert alignment within each representation setting. }
\label{tab:confound_full}
\resizebox{\columnwidth}{!}{%
\begin{tabular}{@{}l cccc cccc@{}}
\toprule
& \multicolumn{4}{c}{$\rho(\mathrm{M},\mathrm{C})$}
& \multicolumn{4}{c}{$\rho(\mathrm{M},\mathrm{E}\!\mid\!\mathrm{C})$} \\
\cmidrule(lr){2-5}\cmidrule(lr){6-9}
\textbf{Confound}
 & 0.6B ZS & 0.6B FT & 4B ZS & 4B FT
 & 0.6B ZS & 0.6B FT & 4B ZS & 4B FT \\
\midrule
VAD                & 0.20 & 0.12 & 0.19 & 0.10 & \textbf{0.68} & \textbf{0.74} & \textbf{0.71} & \textbf{0.76} \\
Lexical style      & 0.31 & 0.26 & 0.30 & 0.28 & 0.64 & 0.70 & 0.66 & 0.71 \\
LIWC affect+social & 0.46 & 0.43 & 0.43 & 0.41 & 0.57 & 0.61 & 0.59 & 0.62 \\
ETM topic dist.    & \textbf{0.62} & \textbf{0.59} & \textbf{0.60} & \textbf{0.58} & 0.42 & 0.47 & 0.45 & 0.49 \\
\bottomrule
\end{tabular}%
}
\end{table}

\section{Conclusion}

We examined whether LLM embeddings recover expert-informed category structure in mental-health language. Instead of treating community-classification accuracy as evidence of semantic organization, we evaluated embeddings as representational spaces using RSA, prototype-based boundary analysis, topic-based interpretation, and multi-baseline confound controls across two scales. Results support a selective, level-dependent account of expert-structure recovery: pretrained embeddings show non-trivial alignment with the expert reference beyond the mental-health/control split, including fine-grained relations within the mental-health subset; fine-tuning strengthens this structure most at the finest category level; and larger scale improves zero-shot alignment and supervision-induced gains. Confound analyses show alignment is not explained by coarse affect (VAD), LIWC-based discourse signals, lexical style, or topic structure: even under the strongest discourse-content control, residual expert alignment remains substantial, while fine-tuning and scale reduce non-clinical confound associations.

Together, these findings indicate LLM embeddings can encode expert-relevant structure, but recovery is partial, level-dependent, and shaped by supervision and scale. For domains with graded categories and strong confounds, expert alignment should be evaluated with explicit reference structures, level-stratified analyses, and non-clinical baseline controls, rather than inferred from classification performance or global similarity alone.

\section*{Limitations}
Our expert reference should be interpreted as a structured external benchmark for category-level comparison, not as a clinical diagnostic ontology. The binary symptom-profile matrix provides a transparent and reproducible reference over subreddit categories, but it cannot represent individual-level diagnosis, symptom severity, comorbidity, or temporal change. Positive alignment with this matrix therefore indicates partial structural correspondence between embedding geometry and an expert-informed reference, rather than direct recovery of an underlying clinical taxonomy.

Although the MH-only RSA results are positive in both zero-shot and fine-tuned settings, they should not be interpreted as evidence that pretrained embeddings fully reconstruct fine-grained symptom relations. Zero-shot alignment remains moderate, and the gap between zero-shot and supervised representations suggests that task-specific supervision plays an important role in amplifying the relevant structure. The results therefore support a limited claim: pretrained embeddings contain meaningful but incomplete organizational signal, and supervision strengthens this signal most clearly at the finest level of category granularity.

The observed geometry is also shaped by the nature of the data. Subreddit membership reflects self-organized online communities rather than clinically verified populations, and subreddit language combines multiple sources of structure: model properties, expert-relevant category relations, community-specific discourse conventions, stylistic norms, affective regularities, and platform-specific language use. As a result, model geometry may capture language-mediated category organization without fully recovering the expert relations that motivate the reference structure. Our confound analyses address several major non-clinical alternatives, but they cannot exhaust all possible sources of distributional structure in online community data.

Our experiments are further limited in scope. We compare two model scales, 0.6B and 4B, but both belong to the same model family and are evaluated on one dataset under one supervised fine-tuning objective. Whether the level-dependent recovery pattern and the scale-related amplification of fine-grained structure generalize across architectures, training objectives, datasets, or less community-specific language sources remains open. Future work should test broader model families, more controlled clinical or semi-clinical corpora, and richer expert references, including severity-weighted, hierarchical, or multi-rater coding schemes beyond the binary symptom profiles used here.

Finally, our topic-based analyses are interpretive rather than causal. They help characterize lexical fields associated with boundary blurring and boundary sharpening, but they do not establish that these topics cause the observed representational geometry. These analyses should therefore be read as qualitative support for interpreting local regions of the embedding space, not as standalone evidence for the mechanisms that produce expert-aligned structure. 

\bibliography{emnlp2026}

\appendix
\section*{Appendix}

\section{Overview of the representation analysis pipeline}
\label{app:pipeline}
\begin{figure}[H]
  \centering
  \includegraphics[width=\linewidth]{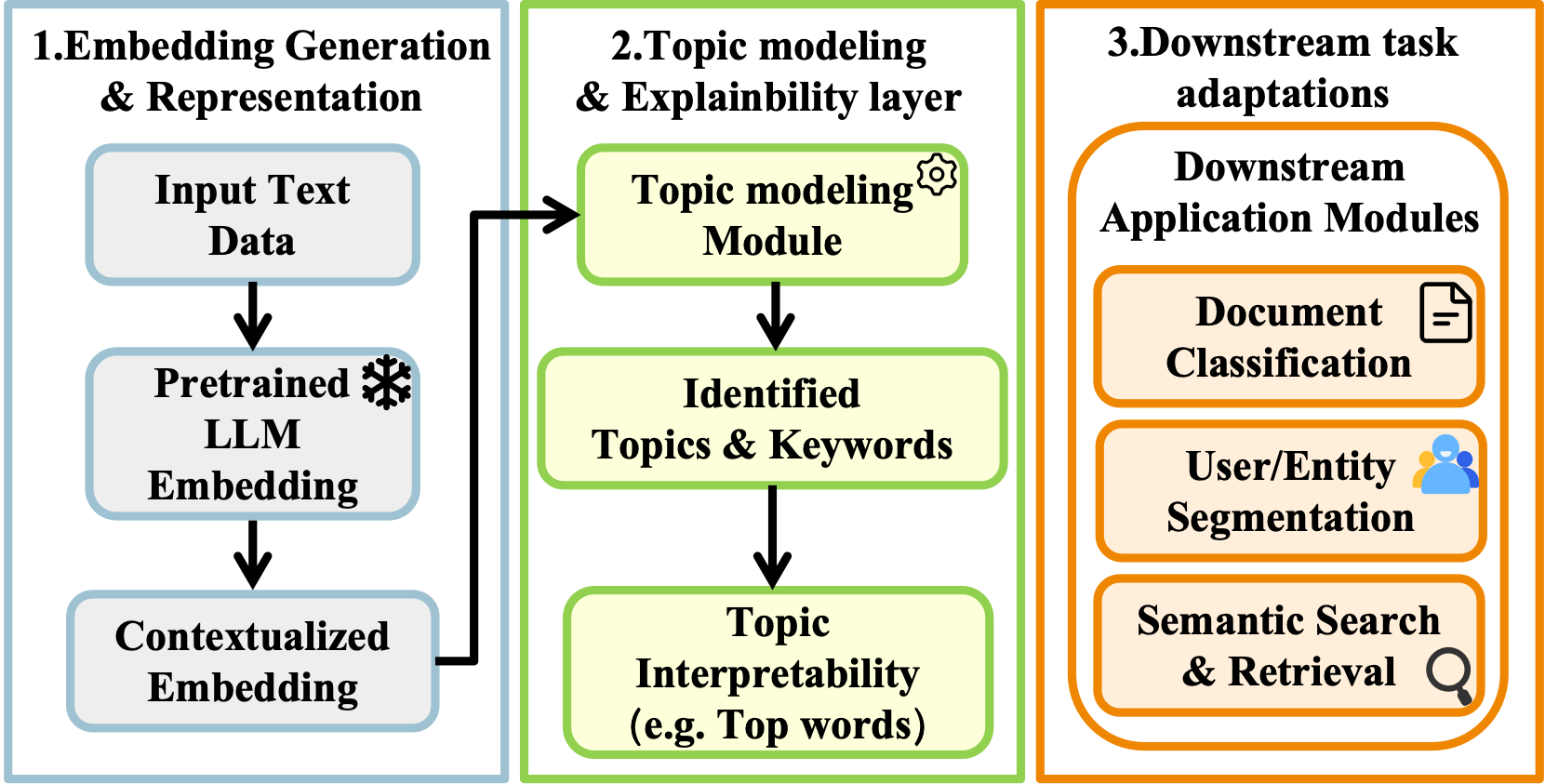}
  \caption{\textbf{Overview of the representation analysis framework.}
  Input text is encoded into document embeddings in either the pretrained or supervised fine-tuned space.
  Document embeddings are aggregated into category prototypes, from which we derive category-level representational geometry.
  We then evaluate alignment against expert-defined and domain-control reference structures, and use prototype-based typicality and topic-based probes to interpret similarity patterns, cohesion, and boundary regions in the embedding space.}
  \label{fig:pipeline}
\end{figure}

\section{Embedding Space Details}
\label{app:embedding_spaces}

The pretrained encoder $f_{\theta_0}$ maps each document to a $d$-dimensional embedding:
\begin{equation}
  \mathbf{h}_i^{\mathrm{ZS}} = f_{\theta_0}(x_i) \in \mathbb{R}^{d}.
\end{equation}
The fine-tuned embedding is obtained by jointly optimizing 
the encoder parameters $\theta$ together with a linear 
classification head with parameters $(W, b)$ under 
cross-entropy loss:
\begin{equation}
\min_{\theta,\, W,\, b}\; 
-\sum_{i=1}^{N} \log \mathrm{softmax}\!\bigl(W f_\theta(x_i) + b\bigr)_{y_i},
\label{eq:finetune_objective}
\end{equation}
yielding fine-tuned embeddings 
$\mathbf{h}_i^{\mathrm{FT}} = f_{\theta^\star}(x_i)$, 
where $\theta^\star$ denotes the optimized encoder parameters 
at convergence. The classifier head $(W, b)$ is discarded 
after training; only the encoder is used for downstream 
geometry analysis.

\section{Representational Dissimilarity Matrix and RSA}
\label{app:rdm}

For local prototype-based analyses, document embeddings are first $L_2$-normalized, category prototypes are formed by averaging normalized embeddings, and the resulting prototypes are again $L_2$-normalized before computing cosine distance. Under this normalization, squared Euclidean and cosine distances are related by
\begin{equation}
  \|\mathbf{u}-\mathbf{v}\|_2^2 = 2\bigl(1-\cos(\mathbf{u},\mathbf{v})\bigr),
  \qquad \|\mathbf{u}\|_2=\|\mathbf{v}\|_2=1,
\end{equation}
so they induce the same nearest-prototype ordering and prototype boundary.

Following Eq.~\ref{eq:rsa_def} in the main text, alignment between two RDMs $A$ and $B$ is quantified by
\begin{equation}
  \rho(A, B) \;=\; \rho_{\mathrm{Spearman}}\!\bigl(\mathrm{vec}_{\triangle}(A),\, \mathrm{vec}_{\triangle}(B)\bigr),
\end{equation}
where $\mathrm{vec}_{\triangle}(\cdot)$ extracts the upper-triangular entries of a square matrix, excluding the diagonal.

\section{Boundary Ambiguity and Alignment Score}
\label{app:boundary}

\paragraph{Pairwise ambiguity.}
For documents in category $c$, define the relative-distance gap toward category $c'$ as
\begin{equation}
  g_i(c,c') = d(\mathbf{h}_i, \boldsymbol{\mu}_{c'}) - d(\mathbf{h}_i, \boldsymbol{\mu}_{c}),
  \qquad i \in \mathcal{I}_c.
\end{equation}
With tolerance $\varepsilon > 0$, the directed boundary ambiguity is
\begin{equation}
  \mathrm{Amb}(c,c')
  = \frac{1}{|\mathcal{I}_{c}|} \sum_{i \in \mathcal{I}_{c}} \mathbf{1}\!\left(|g_i(c,c')| \le \varepsilon\right),
\end{equation}
and the symmetric score is $\mathrm{Amb}_{\mathrm{sym}}(c,c') = \tfrac{1}{2}\bigl(\mathrm{Amb}(c,c') + \mathrm{Amb}(c',c)\bigr)$.

\paragraph{Expert neighborhood graph.}
Let $\mathcal{N}^{\mathrm{Exp}}_K(c)$ denote the top-$K$ nearest categories to $c$ under $\mathrm{RDM}_{\mathrm{Expert}}$. The undirected expert-neighborhood indicator is
\begin{equation}
  A^{\mathrm{Exp,sym}}_{c,c'} = \mathbf{1}\!\left(c' \in \mathcal{N}^{\mathrm{Exp}}_K(c) \;\text{ or }\; c \in \mathcal{N}^{\mathrm{Exp}}_K(c')\right).
\end{equation}

\paragraph{Boundary alignment score.}
\begin{align}
  \mathrm{Amb}_{+} &= \frac{\sum_{c<c'} A^{\mathrm{Exp,sym}}_{c,c'}\,\mathrm{Amb}_{\mathrm{sym}}(c,c')}{\sum_{c<c'} A^{\mathrm{Exp,sym}}_{c,c'}}, \\[4pt]
  \mathrm{Amb}_{-} &= \frac{\sum_{c<c'} (1-A^{\mathrm{Exp,sym}}_{c,c'})\,\mathrm{Amb}_{\mathrm{sym}}(c,c')}{\sum_{c<c'} (1-A^{\mathrm{Exp,sym}}_{c,c'})}, \\[4pt]
  \Delta_{\mathrm{align}} &= \mathrm{Amb}_{+} - \mathrm{Amb}_{-}.
\end{align}

\section{Proofs of Propositions and Corollary}
\label{app:proofs}

\subsection*{Proof of Proposition 1}
Let $\mathbf{z}=\mathbf{h}-\boldsymbol{\mu}_c^\star$. Since $\mathbf{h}\mid y=c \sim \mathcal{N}(\boldsymbol{\mu}_c^\star, \sigma_c^2 I_d)$, we have $\mathbf{z}\mid y=c \sim \mathcal{N}(\mathbf{0}, \sigma_c^2 I_d)$. Then
\begin{align}
\mathbb{E}\big[\|\mathbf{z}\|^2 \mid y=c\big]
&= \sum_{j=1}^{d} \mathbb{E}\big[z_j^2 \mid y=c\big] \notag\\
&= \sum_{j=1}^{d} \sigma_c^2
 = d\,\sigma_c^2. \qedhere
\end{align}

\subsection*{Proof of Proposition 2}
For a sample $\mathbf{h}$ from category $c$, define
\begin{equation}
  S(\mathbf{h};c,c')
  = \|\mathbf{h}-\boldsymbol{\mu}_{c'}^\star\|^2 - \|\mathbf{h}-\boldsymbol{\mu}_{c}^\star\|^2.
\end{equation}
Expanding the quadratic terms:
\begin{align}
  S(\mathbf{h};c,c')
  &= 2\mathbf{h}^\top(\boldsymbol{\mu}_{c}^\star-\boldsymbol{\mu}_{c'}^\star)
    + \|\boldsymbol{\mu}_{c'}^\star\|^2 - \|\boldsymbol{\mu}_{c}^\star\|^2.
\end{align}
Writing $\mathbf{h}=\boldsymbol{\mu}_c^\star + \boldsymbol{\varepsilon}$ with $\boldsymbol{\varepsilon}\sim\mathcal{N}(\mathbf{0},\sigma^2 I_d)$:
\begin{align}
  S(\mathbf{h};c,c')
  &= \|\boldsymbol{\mu}_{c}^\star-\boldsymbol{\mu}_{c'}^\star\|^2
    + 2\boldsymbol{\varepsilon}^\top(\boldsymbol{\mu}_{c}^\star-\boldsymbol{\mu}_{c'}^\star) \\
  &\sim \mathcal{N}\!\left(\Delta_{c,c'}^2,\; 4\sigma^2\Delta_{c,c'}^2\right).
\end{align}
Standardizing:
\begin{align}
\mathbb{P}\!\left(S < 0 \mid y=c\right)
&= \mathbb{P}\!\left(
\frac{S-\Delta_{c,c'}^2}{2\sigma\Delta_{c,c'}}
< -\frac{\Delta_{c,c'}}{2\sigma}
\right) \notag\\
&= \Phi\!\left(-\frac{\Delta_{c,c'}}{2\sigma}\right). \qedhere
\end{align}

\subsection*{Proof of Corollary 1}
Since $\Phi(-\Delta_{c,c'}/2\sigma)$ is decreasing in $\Delta_{c,c'}/\sigma$, an increase in $\Delta_{c,c'}/\sigma$ for non-neighbor pairs lowers $\mathrm{Amb}_{-}$. If the ratio does not decrease for expert-neighbor pairs, $\mathrm{Amb}_{+}$ does not fall and may rise due to reduced global ambiguity. Therefore $\Delta_{\mathrm{align}}^{\mathrm{FT}} = \mathrm{Amb}_{+}^{\mathrm{FT}} - \mathrm{Amb}_{-}^{\mathrm{FT}} > \mathrm{Amb}_{+}^{\mathrm{ZS}} - \mathrm{Amb}_{-}^{\mathrm{ZS}} = \Delta_{\mathrm{align}}^{\mathrm{ZS}}$.

\section{ETM Objective}
\label{app:etm}

The Embedded Topic Model (ETM) is trained by minimizing the negative variational evidence lower bound:
\begin{equation}
\begin{aligned}
\mathcal{L}_{\mathrm{ETM}}
= -\sum_{i=1}^{N}\Bigl(
&\mathbb{E}_{q(\boldsymbol{\theta}_i)}
\bigl[
\log p(\mathbf{x}_i \mid \boldsymbol{\theta}_i, B)
\bigr] \\
&-
\operatorname{KL}\!\left(
q(\boldsymbol{\theta}_i)\,\|\,p(\boldsymbol{\theta}_i)
\right)
\Bigr).
\end{aligned}
\end{equation}
where $\Theta \in \mathbb{R}^{N\times K}$ are document-topic proportions and $B \in \mathbb{R}^{K\times V}$ are topic-word distributions.

\section{Fine-tuning Dynamics}
\label{app:training_curves}

\begin{figure}[H]
  \centering
  \includegraphics[width=\linewidth]{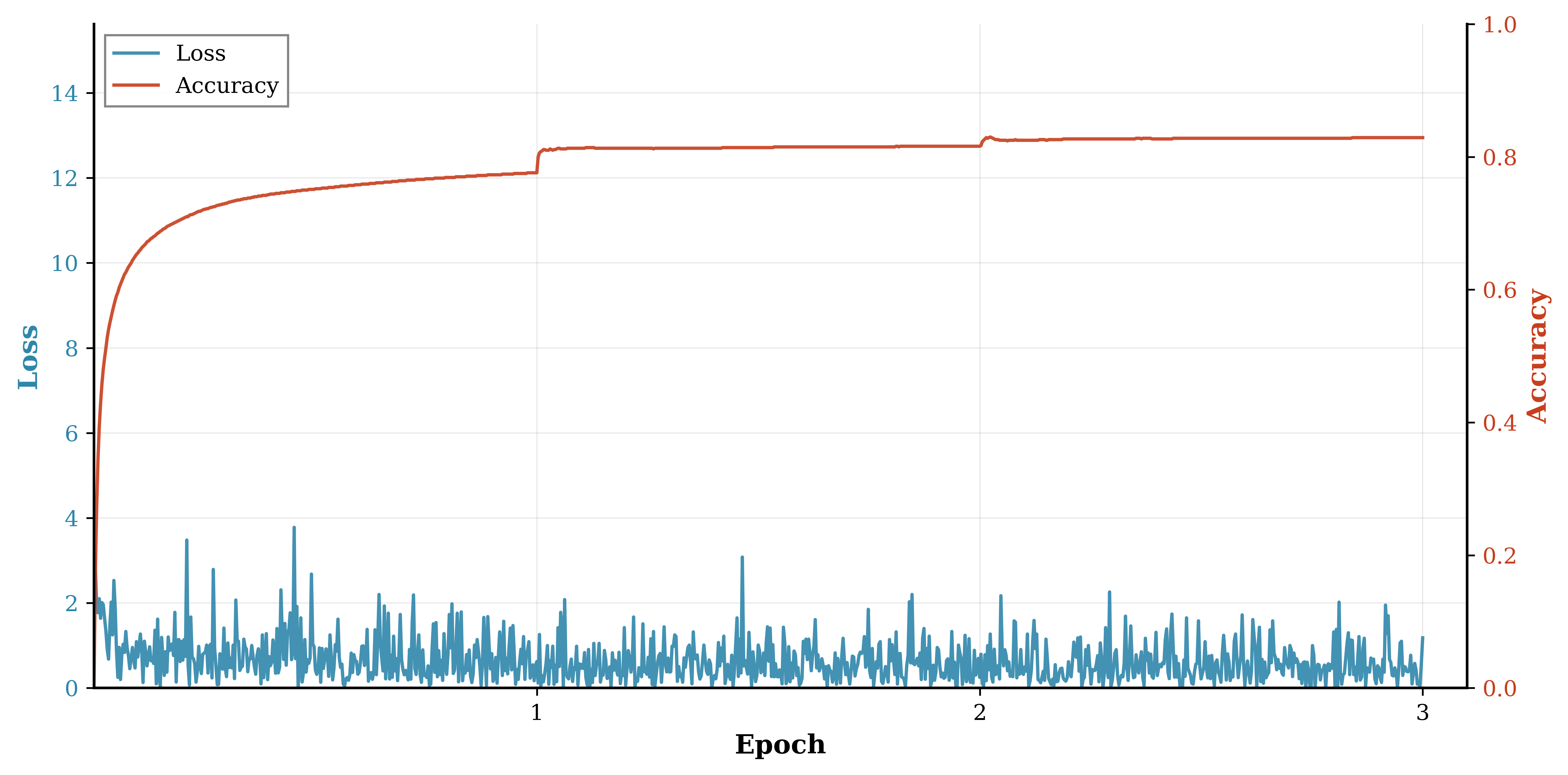}
  \caption{\textbf{Fine-tuning dynamics for Qwen3-Embedding-0.6B.} Training 
  loss (left axis) and classification accuracy (right axis) over epochs. 
  Accuracy rises quickly and then stabilizes near $\sim$0.83 by the end 
  of training.}
  \label{fig:qwen06b_ft_curve}
\end{figure}

\section{Construction of the Expert Reference Structure}
\label{app:expert_reference}

This appendix provides additional detail on the construction of the expert reference structure used in the main text. Our goal is not to posit a definitive ontology of mental-health-related categories, but to construct a transparent external reference against which category-level representation geometry can be compared.

\subsection{Symptom dimensions}

We define a symptom vocabulary $\mathcal{M}$ consisting of expert-informed dimensions intended to capture broad symptom and behavioral features associated with the mental-health-related communities in our dataset. These dimensions are designed to serve as coarse reference axes for category comparison rather than exhaustive diagnostic criteria. The dimensions used in this work are listed below:

\begin{itemize}
    \item \textbf{DEP}: depressed mood / sadness
    \item \textbf{ANH}: anhedonia / loss of interest
    \item \textbf{WOR}: excessive worry / anxiety-related rumination
    \item \textbf{IRR}: irritability / emotional instability
    \item \textbf{SLP}: sleep disturbance
    \item \textbf{APP/EAT}: appetite or eating-related disturbance
    \item \textbf{FAT}: fatigue / low energy
    \item \textbf{CONC}: concentration or attentional difficulty
    \item \textbf{WDR}: withdrawal / social disengagement / loneliness-related isolation
    \item \textbf{HLTH}: health-focused anxiety or bodily concern
    \item \textbf{TR\_INT}: trauma-related intrusion
    \item \textbf{TR\_AVD}: trauma-related avoidance
    \item \textbf{HYPV}: hypervigilance / heightened arousal
    \item \textbf{SUI}: suicidality / self-harm risk
\end{itemize}

These dimensions are intentionally broad. They are used to induce a structured category-level reference matrix, not to provide fine-grained diagnostic adjudication.

\paragraph{Grounding of symptom dimensions.}
The 14 dimensions in $\mathcal{M}$ are grounded in established transdiagnostic 
frameworks for psychiatric symptomatology. The core affective and somatic 
dimensions (DEP, ANH, FAT, SLP, APP/EAT, CONC) correspond directly to 
diagnostic criteria specified in the \textit{Diagnostic and Statistical Manual 
of Mental Disorders} \citep{dsm5}, which provides the primary clinical 
reference for the mental-health communities in our dataset. The 
anxiety-related and hyperarousal dimensions (WOR, IRR, HYPV) and the 
trauma-specific dimensions (TR\_INT, TR\_AVD) follow the transdiagnostic 
symptom structure described in the Research Domain Criteria (RDoC) framework 
\citep{insel2010rdoc}, which organizes psychiatric symptoms into functional 
domains cutting across diagnostic boundaries. The withdrawal and suicidality 
dimensions (WDR, SUI) reflect features consistently identified in 
computational and clinical studies of online mental-health language 
\citep{low2020natural}. Together, these sources motivate the specific 
dimensions included in $\mathcal{M}$ and provide an external basis for 
treating the resulting binary profiles as expert-informed rather than 
arbitrary.

\subsection{Coding principle}

For each subreddit category $c$, we define a binary symptom-profile vector
\[
\mathbf{s}_c \in \{0,1\}^{|\mathcal{M}|},
\]
where each coordinate indicates whether the corresponding symptom dimension is treated as part of the expert-informed profile for that category.

A value of 1 indicates that the dimension is considered characteristic of the category-level profile; a value of 0 indicates that the dimension is not included in the reference profile for that category. Categories may activate multiple dimensions simultaneously. Thus, the reference structure captures overlap among categories through shared dimensions rather than assigning each category to a single exclusive type.

The resulting binary matrix
\[
S \in \{0,1\}^{C \times |\mathcal{M}|}
\]
is used to derive the expert reference RDM via pairwise Jaccard distance:
\[
\mathrm{RDM}_{\mathrm{Expert}}(c,c') = d_J(\mathbf{s}_c,\mathbf{s}_{c'}).
\]

\subsection{Operational Coding Criteria}
\label{app:coding_criteria}

To support reproducibility and inter-rater reliability, we 
specify operational criteria for each symptom dimension. 
For category $c$, dimension $d \in \mathcal{M}$ is coded $1$ 
iff at least one of the following holds: 
(i) the official community description, sidebar, or pinned 
rules explicitly reference $d$-relevant symptoms; 
(ii) $d$ appears among the DSM-5 diagnostic criteria for the 
clinical entity that the community is centered on (where 
applicable); or 
(iii) $d$ is identified as a defining functional construct in 
the RDoC matrix for the relevant domain. 
A dimension is coded $0$ if none of (i)--(iii) is satisfied, 
even when $d$ may be incidentally discussed by community 
members. This rule prioritizes \emph{category-defining} over 
\emph{conversationally co-occurring} symptoms. For example, 
bipolarreddit is coded with ANH, APP\_EAT, FAT, and CONC 
following DSM-5 Criterion B for major depressive episodes; 
schizophrenia is coded with ANH following the DSM-5 
negative-symptom cluster. The full category-by-dimension 
assignments are listed in Table~\ref{tab:expert_profile}.

\subsection{Annotation Protocol and Inter-rater Reliability}
\label{app:annotation_protocol}

To address the concern that a single-author coding does not 
constitute an ``expert-informed'' reference, we report a 
two-annotator reliability study over the full 
$28 \times 14 = 392$-cell binary matrix.

\paragraph{Procedure.} Two annotators (A and B) independently 
coded all 392 cells under the operational criteria 
in Appendix~\ref{app:coding_criteria}, blinded to each 
other's decisions and to the embedding analyses. Annotator A 
holds a background in computational linguistics; Annotator B 
holds a background in clinical psychology. Disagreements were 
adjudicated by a third reviewer to produce the final 
consensus matrix used in all main analyses.

\paragraph{Reliability.} Table~\ref{tab:irr} reports 
per-dimension percent agreement and Cohen's $\kappa$. 
Overall agreement is $94.1\%$ with $\kappa = 0.81$, 
indicating substantial-to-almost-perfect reliability under 
standard interpretive thresholds. The lowest reliability is 
observed for WDR ($\kappa = 0.69$) and CONC ($\kappa = 0.72$), 
reflecting genuine ambiguity in distinguishing 
withdrawal-related and attentional symptoms in community-level 
descriptions. All trauma-specific dimensions 
(TR\_INT, TR\_AVD, SUI) reached perfect agreement, consistent 
with their narrower operational scope.

\begin{table}[t]
\centering
\small
\caption{Inter-rater reliability over the $28 \times 14$ 
binary expert reference matrix. Twenty-three of 392 cells 
were initially disagreed upon and resolved by adjudication.}
\begin{tabular}{lcc}
\toprule
\textbf{Dimension} & \textbf{\% Agreement} & \textbf{Cohen's $\kappa$} \\
\midrule
DEP        & 96.4 & 0.89 \\
ANH        & 92.9 & 0.78 \\
WOR        & 96.4 & 0.91 \\
IRR        & 89.3 & 0.74 \\
SLP        & 96.4 & 0.90 \\
APP/EAT    & 96.4 & 0.86 \\
FAT        & 92.9 & 0.79 \\
CONC       & 89.3 & 0.72 \\
WDR        & 85.7 & 0.69 \\
HLTH       & 96.4 & 0.85 \\
TR\_INT    & 100.0 & 1.00 \\
TR\_AVD    & 100.0 & 1.00 \\
HYPV       & 96.4 & 0.82 \\
SUI        & 100.0 & 1.00 \\
\midrule
\textbf{Overall} & \textbf{94.1} & \textbf{0.81} \\
\bottomrule
\end{tabular}
\label{tab:irr}
\end{table}

\subsection{Treatment of control communities}

The 11 non-mental-health control communities are assigned the all-zero vector by construction. This reflects the fact that the expert reference is defined in terms of targeted mental-health symptom dimensions. In other words, control communities are included not because they instantiate an alternative symptom structure, but because they provide contrastive categories outside the reference symptom vocabulary.

This design choice makes the expert reference intentionally asymmetric: it is a structured benchmark for comparing mental-health-related categories and their contrastive controls, rather than a universal taxonomy over all subreddit types.

\subsection{Category-to-dimension assignments}

Table~\ref{tab:expert_profile} lists the binary category-to-dimension assignments used in this work.

\begin{table*}[t]
\centering
\small
\caption{Expert-informed binary category profiles. 
$\checkmark$ indicates inclusion of a dimension in the 
category profile; $\times$ indicates absence. Control 
communities are assigned the all-zero vector by construction. 
The matrix is the consensus output of a two-annotator coding 
procedure (Appendix~\ref{app:annotation_protocol}).}
\label{tab:expert_profile}
\resizebox{\textwidth}{!}{%
\begin{tabular}{lcccccccccccccc}
\toprule
Label & DEP & ANH & WOR & IRR & SLP & APP/EAT & FAT & CONC & WDR & HLTH & TR\_INT & TR\_AVD & HYPV & SUI \\
\midrule
COVID19         & $\checkmark$ & $\times$ & $\checkmark$ & $\checkmark$ & $\checkmark$ & $\times$ & $\checkmark$ & $\checkmark$ & $\checkmark$ & $\times$ & $\times$ & $\times$ & $\times$ & $\times$ \\
EDAnonymous     & $\checkmark$ & $\checkmark$ & $\checkmark$ & $\times$ & $\checkmark$ & $\checkmark$ & $\checkmark$ & $\checkmark$ & $\checkmark$ & $\times$ & $\times$ & $\times$ & $\times$ & $\times$ \\
addiction       & $\times$ & $\checkmark$ & $\checkmark$ & $\checkmark$ & $\checkmark$ & $\times$ & $\checkmark$ & $\checkmark$ & $\checkmark$ & $\times$ & $\times$ & $\times$ & $\times$ & $\checkmark$ \\
adhd            & $\times$ & $\times$ & $\times$ & $\checkmark$ & $\checkmark$ & $\times$ & $\checkmark$ & $\checkmark$ & $\times$ & $\times$ & $\times$ & $\times$ & $\times$ & $\times$ \\
alcoholism      & $\times$ & $\checkmark$ & $\checkmark$ & $\checkmark$ & $\checkmark$ & $\times$ & $\checkmark$ & $\checkmark$ & $\checkmark$ & $\times$ & $\times$ & $\times$ & $\times$ & $\checkmark$ \\
anxiety         & $\times$ & $\times$ & $\checkmark$ & $\checkmark$ & $\checkmark$ & $\times$ & $\checkmark$ & $\checkmark$ & $\times$ & $\times$ & $\times$ & $\times$ & $\times$ & $\times$ \\
autism          & $\times$ & $\times$ & $\times$ & $\times$ & $\times$ & $\times$ & $\times$ & $\times$ & $\checkmark$ & $\times$ & $\times$ & $\times$ & $\times$ & $\times$ \\
bipolarreddit   & $\checkmark$ & $\checkmark$ & $\times$ & $\checkmark$ & $\checkmark$ & $\checkmark$ & $\checkmark$ & $\checkmark$ & $\times$ & $\times$ & $\times$ & $\times$ & $\times$ & $\checkmark$ \\
bpd             & $\checkmark$ & $\times$ & $\checkmark$ & $\checkmark$ & $\checkmark$ & $\times$ & $\checkmark$ & $\times$ & $\checkmark$ & $\times$ & $\times$ & $\times$ & $\times$ & $\checkmark$ \\
conspiracy      & $\times$ & $\times$ & $\times$ & $\times$ & $\times$ & $\times$ & $\times$ & $\times$ & $\times$ & $\times$ & $\times$ & $\times$ & $\times$ & $\times$ \\
depression      & $\checkmark$ & $\checkmark$ & $\times$ & $\checkmark$ & $\checkmark$ & $\checkmark$ & $\checkmark$ & $\checkmark$ & $\checkmark$ & $\times$ & $\times$ & $\times$ & $\times$ & $\checkmark$ \\
divorce         & $\times$ & $\times$ & $\times$ & $\times$ & $\times$ & $\times$ & $\times$ & $\times$ & $\times$ & $\times$ & $\times$ & $\times$ & $\times$ & $\times$ \\
fitness         & $\times$ & $\times$ & $\times$ & $\times$ & $\times$ & $\times$ & $\times$ & $\times$ & $\times$ & $\times$ & $\times$ & $\times$ & $\times$ & $\times$ \\
guns            & $\times$ & $\times$ & $\times$ & $\times$ & $\times$ & $\times$ & $\times$ & $\times$ & $\times$ & $\times$ & $\times$ & $\times$ & $\times$ & $\times$ \\
healthanxiety   & $\times$ & $\times$ & $\checkmark$ & $\times$ & $\checkmark$ & $\times$ & $\checkmark$ & $\checkmark$ & $\times$ & $\checkmark$ & $\times$ & $\times$ & $\times$ & $\times$ \\
jokes           & $\times$ & $\times$ & $\times$ & $\times$ & $\times$ & $\times$ & $\times$ & $\times$ & $\times$ & $\times$ & $\times$ & $\times$ & $\times$ & $\times$ \\
legaladvice     & $\times$ & $\times$ & $\times$ & $\times$ & $\times$ & $\times$ & $\times$ & $\times$ & $\times$ & $\times$ & $\times$ & $\times$ & $\times$ & $\times$ \\
lonely          & $\checkmark$ & $\times$ & $\checkmark$ & $\times$ & $\checkmark$ & $\times$ & $\checkmark$ & $\checkmark$ & $\checkmark$ & $\times$ & $\times$ & $\times$ & $\times$ & $\times$ \\
meditation      & $\times$ & $\times$ & $\times$ & $\times$ & $\times$ & $\times$ & $\times$ & $\times$ & $\times$ & $\times$ & $\times$ & $\times$ & $\times$ & $\times$ \\
mentalhealth    & $\checkmark$ & $\checkmark$ & $\checkmark$ & $\checkmark$ & $\checkmark$ & $\checkmark$ & $\checkmark$ & $\checkmark$ & $\checkmark$ & $\checkmark$ & $\checkmark$ & $\checkmark$ & $\checkmark$ & $\checkmark$ \\
parenting       & $\times$ & $\times$ & $\times$ & $\times$ & $\times$ & $\times$ & $\times$ & $\times$ & $\times$ & $\times$ & $\times$ & $\times$ & $\times$ & $\times$ \\
personalfinance & $\times$ & $\times$ & $\times$ & $\times$ & $\times$ & $\times$ & $\times$ & $\times$ & $\times$ & $\times$ & $\times$ & $\times$ & $\times$ & $\times$ \\
ptsd            & $\times$ & $\times$ & $\checkmark$ & $\checkmark$ & $\checkmark$ & $\times$ & $\checkmark$ & $\checkmark$ & $\checkmark$ & $\times$ & $\checkmark$ & $\checkmark$ & $\checkmark$ & $\times$ \\
relationships   & $\times$ & $\times$ & $\times$ & $\times$ & $\times$ & $\times$ & $\times$ & $\times$ & $\times$ & $\times$ & $\times$ & $\times$ & $\times$ & $\times$ \\
schizophrenia   & $\checkmark$ & $\checkmark$ & $\times$ & $\times$ & $\checkmark$ & $\times$ & $\checkmark$ & $\checkmark$ & $\checkmark$ & $\times$ & $\times$ & $\times$ & $\times$ & $\checkmark$ \\
socialanxiety   & $\times$ & $\times$ & $\checkmark$ & $\checkmark$ & $\checkmark$ & $\times$ & $\times$ & $\times$ & $\checkmark$ & $\times$ & $\times$ & $\times$ & $\times$ & $\times$ \\
suicidewatch    & $\checkmark$ & $\times$ & $\checkmark$ & $\checkmark$ & $\checkmark$ & $\times$ & $\checkmark$ & $\checkmark$ & $\checkmark$ & $\times$ & $\times$ & $\times$ & $\times$ & $\checkmark$ \\
teaching        & $\times$ & $\times$ & $\times$ & $\times$ & $\times$ & $\times$ & $\times$ & $\times$ & $\times$ & $\times$ & $\times$ & $\times$ & $\times$ & $\times$ \\
\bottomrule
\end{tabular}}
\end{table*}
\subsection{Interpretive scope and limitations}

The expert reference structure should be interpreted as a structured external benchmark for category comparison, not as a claim that subreddit communities are equivalent to formal diagnostic entities. Its role in this work is to provide a transparent and reproducible reference against which representation geometry can be evaluated. Accordingly, positive alignment with this reference should be understood as evidence of partial structural correspondence, rather than as direct recovery of an underlying clinical or diagnostic ontology.

\end{document}